\documentclass[11pt]{article}
\usepackage{acl}
\usepackage{times}
\usepackage{amsmath}
\usepackage{amssymb}
\usepackage{latexsym}
\usepackage{float}
\usepackage{multirow}
\usepackage{tabularx}
\usepackage{graphicx} 
\usepackage{inconsolata}
\usepackage[utf8]{inputenc}
\usepackage{microtype}
\usepackage[T1]{fontenc}
\usepackage{listings}
\usepackage{booktabs}
\usepackage{array}
\title{Question-to-Question Retrieval for Hallucination-Free Knowledge Access: An Approach for Wikipedia and Wikidata Question Answering}
\author{Santhosh Thottingal \\
\texttt{santhosh.thottingal@gmail.com}
}
\date{February 2025}

\begin{document}

\maketitle

\begin{abstract}
This paper introduces an approach to question answering over knowledge bases like Wikipedia and Wikidata by performing "question-to-question" matching and retrieval from a dense vector embedding store. Instead of embedding document content, we generate a comprehensive set of questions for each logical content unit using an instruction-tuned LLM. These questions are vector-embedded and stored, mapping to the corresponding content. Vector embedding of user queries are then matched against this question vector store. The highest similarity score leads to direct retrieval of the associated article content, eliminating the need for answer generation. Our method achieves high cosine similarity ( > 0.9 ) for relevant question pairs, enabling highly precise retrieval. This approach offers several advantages including computational efficiency, rapid response times, and increased scalability. We demonstrate its effectiveness on Wikipedia and Wikidata, including multimedia content through structured fact retrieval from Wikidata, opening up new pathways for multimodal question answering.
\end{abstract}
\begin{figure*}
    \centering
    \includegraphics[width=1\textwidth]{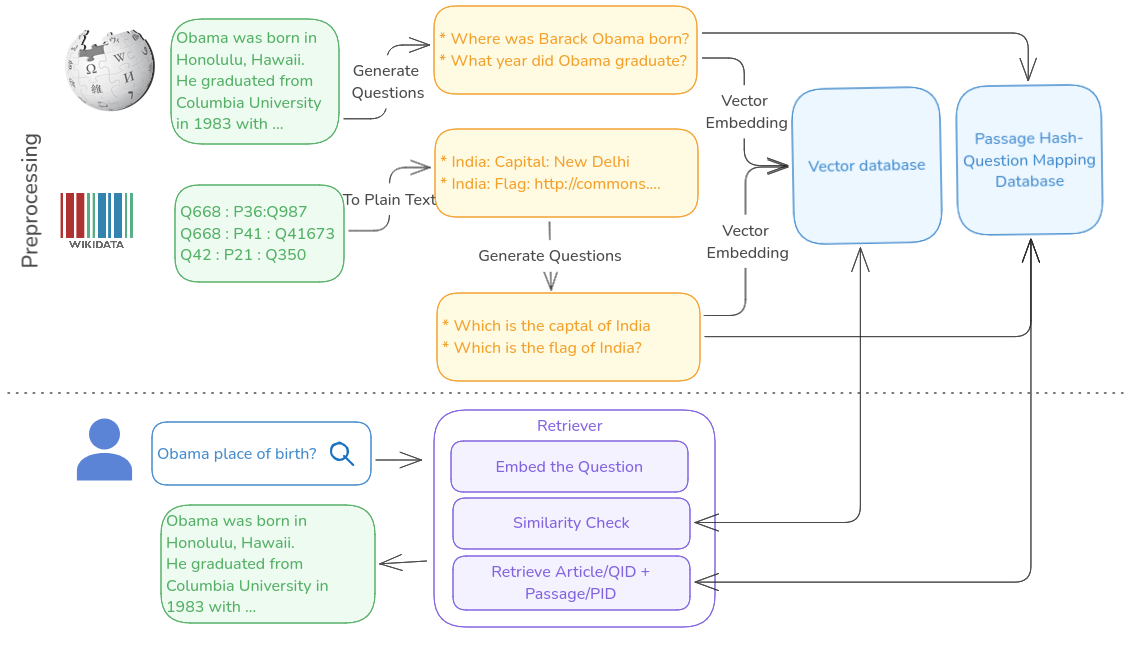}
    \caption{Overall Architecture of Question-Question retrieval and question answering}
    \label{fig:oveview}
\end{figure*}

\section{Introduction}
Question answering (QA) is a task that answers factoid questions using a large collection of documents. In the context of Wikipedia, it is to answer questions beyond the keyword based traditional search. The rise of large language models (LLMs) has opened up new possibilities for building more capable QA systems, yet the challenge remains in ensuring their reliability and avoiding the generation of fabricated or hallucinated responses\cite{gao2024retrievalaugmentedgenerationlargelanguage}. As trustable encylopedic information is the unique value that Wikipedia provides, providing information as accurate as possible is in its mission.

Retrieval-Augmented Generation (RAG) has become a common approach to address this challenge by combining the knowledge retrieval capabilities with the generative power of LLMs \cite{lewis2021retrievalaugmentedgenerationknowledgeintensivenlp}. A significant limitation in conventional RAG models stems from the typically low semantic similarity between vector embeddings of natural language questions and typical document passages\cite{karpukhin-etal-2020-dense}. This disparity arises primarily from their distinct structural differences: questions are interrogative, while passages are predominantly declarative. For example, when querying with "Where is Eiffel Tower located?", the question's embedding is compared against passages like "Eiffel Tower is located in Paris." Although "Paris" is semantically crucial, the contrasting sentence structures can lead to low cosine similarity scores, often falling within the range of 0.4-0.7, even for relevant content. This "question-to-passage" comparison, exacerbated by lexical differences and the differing focus of questions and passages, hinders high-precision retrieval, particularly for queries seeking specific information. Addressing this challenge requires strategies like question reformulation, passage summarization, or hybrid approaches that combine semantic search with keyword matching.

Furthermore, while LLMs excel at generating human-like text, the generation step itself introduces the risk of hallucination, where LLMs may create plausible but factually incorrect answers. We argue that a retrieval mechanism that can precisely identify the relevant text and thereby removing the generation step to mitigate hallucination is a practical and useful approach in the context of Wikipedia.

This paper introduces a novel approach to knowledge-based question answering that addresses these limitations by performing "question-to-question" retrieval and avoids the generation step. Instead of embedding document content, we employ an instruction-tuned LLM to generate a comprehensive set of questions for each logical content unit within the knowledge base. These generated questions are then vector-embedded and stored in a searchable vector store, mapped to the corresponding source document. When a user submits a query, it is also embedded in the same space and a search performed to identify the most similar generated question. For highest matching question, we present the corresponding article and the paragraph with in the article. A similar approach is presented for wikidata where the content is in the form of entity relations, we generate all possible questions and then make the wikidata knowledge graph accessible to a question answer system.

\section{Background}
 
The problem of Wikipedia question answering is as follows: Given a factoid question like "Who is the architect of Eiffel Tower?", "How many people died in Chernobil accident?", the retrieval system takes the user to a Wikipedia article's paragraph that answer's the question. Compared to common RAG techniques, there is no LLM based articulation of answer. We assume the question is extractive, in which the answer is available within such a paragraph and does not need to analyze content across multiple paragraphs or articles.

Assume that our collection contains \(D\) documents, $d_1, d_2,\cdots,d_D$. We first
split each of the Wikipedia article into text passages as the basic retrieval units and get $M$
total passages in our corpus $C = {p_1, p_2,\cdots, p_M }$, 
where each passage $p_i$ can be viewed as a sequence of tokens $w^{(i)}_1 , w^{(i)}_2 , \cdots , w^{(i)}_{|p_i|}$  Given a question $q$, the task is to find a $p$ that can answer the question. For the sake of quick and precise mapping between passage and article, assume there is a unique cryptographic hash based on the content of each $p$ that is mapped to a $d$ and saved in a relational database.

A QA system needs to include an efficient retriever component that can select a small set of relevant texts\cite{chen-etal-2017-reading}.
Formally speaking, a retriever $R : (q, \mathcal{C}) \rightarrow \mathcal{C}_\mathcal{F}$ is a function that takes as input a question $q$ and a corpus $C$ and returns a much smaller filter set of texts $C_\mathcal{F} \subset \mathcal{C}$, where $|\mathcal{C}_\mathcal{F} | = k \ll |\mathcal{C}|$. For a fixed $k$, a retriever can be evaluated in isolation on \textit{top-k} retrieval accuracy, which is the fraction of questions for which $C_\mathcal{F}$ contains a span that answers the question\cite{karpukhin-etal-2020-dense}.

In usual RAG approach, the retrieval process begins with a user's query being converted into a vector representation using a text embedding model. The knowledge base (e.g., a collection of documents or articles) is preprocessed by chunking large bodies of text into smaller, semantically meaningful units such as paragraphs or sentences. These chunks are also embedded into a vector space, often using the same embedding model that was used to embed the user query. The embeddings of these knowledge base chunks are then stored in a vector database or index to enable fast similarity search. During query time, the embedded user query is compared to each chunk vector and similarity metrics such as cosine similarity are used to identify the most relevant chunks from the knowledge base.

A core problem, as highlighted in the introduction, arises from the inherent structural difference between a user's question and a standard document passage. The closest match is not a plausible answer to our question — instead, it is another question\footnote{Don't use cosine similarity carelessly - Piotr Migdał \url{https://p.migdal.pl/blog/2025/01/dont-use-cosine-similarity}}. When a question vector is compared to a passage vector, the comparison relies on shared keywords and latent semantics. However, the semantics of the answer is what defines its usefulness, and comparing it with question embeddings is inherently challenging. This misalignment between the question and passage vector space often leads to low cosine similarity scores even when the retrieved passage contains relevant information. Additionally, the learned embeddings of passages have a degree of freedom that can render arbitrary cosine-similarities\cite{Steck2024}. So,  "question-to-passage" vector comparison can make it hard to reliably retrieve the most appropriate content, potentially diminishing the overall performance of the QA system. As illustrated in table \ref{similarity-score}, for a simple paragraph and questions from it, the usual retrieval score falls below 0.7 and our objective is to maximize that score above 0.9.

\begin{table*}[h!]
\centering
\begin{tabularx}{\linewidth}{ll}
\hline
 \textbf{Question} & \textbf{Similarity Score}\\
 \hline
Where was Barack Obama born?	& 0.68\\ 
Which university did Obama graduate from?	& 0.71\\ 
What year did Obama graduate?	& 0.70\\ 
Where did Obama work as a community organizer?	& 0.57\\ 
Who is the first black president of Harvard Law Review?	& 0.55\\ 
From what years did Obama teach at University of Chicago Law School?	& 0.75\\ 
When was Obama first elected to the Illinois Senate?	& 0.67\\ 
When did Obama run for U.S. Senate? &	0.66\\ 
Who was Obama's running mate in the 2008 presidential election?	& 0.63\\ 
Who did Obama defeat in the 2008 presidential election?	& 0.58\\ 
Who defeated John McCain in the 2008 presidential election?	& 0.63\\ 
What political party nominated Obama for president?	& 0.58\\ 
Obama birth place	& 0.64\\ 
\end{tabularx}
\caption{\label{similarity-score} Questions and similarity score for the following passage:
\textit{Obama was born in Honolulu, Hawaii. 
He graduated from Columbia University in 1983 with a Bachelor of Arts degree in political science and later worked as a community organizer in Chicago.
In 1988, Obama enrolled in Harvard Law School, where he was the first black president of the Harvard Law Review.
He became a civil rights attorney and an academic, teaching constitutional law at the University of Chicago Law School from 1992 to 2004.
In 1996, Obama was elected to represent the 13th district in the Illinois Senate, a position he held until 2004, when he successfully ran for the U.S. Senate.
In the 2008 presidential election, after a close primary campaign against Hillary Clinton, he was nominated by the Democratic Party for president.
Obama selected Joe Biden as his running mate and defeated Republican nominee John McCain.}
Embedding model used: text-embedding-004. Dimensions: 798
}
\end{table*}

Question answering over structured knowledge bases like Wikidata\cite{dennywikidata}, often involves translating natural language queries into structured queries (e.g., SPARQL) and executing these queries against the KB. Approaches vary from using rule-based systems to more sophisticated neural network models that learn to map natural language to formal query representations\cite{liu2024spinachsparqlbasedinformationnavigation}. These techniques heavily rely on the structural aspects of the knowledge base, which is different from our approach. We aim to use the structured data in wikidata as knowledge based for factoid questions without using SPARQL queries.

\section{Methodology}

We focus our research in this work on improving the retrieval component in open-domain QA.
Given a collection of $M$ text passages, the goal of our retriever is to index all the passages in a low-dimensional and continuous space, such that it can retrieve efficiently the top $k$ passages relevant to the input question for the reader at run-time. Note that $M$ can be very large(e.g., 6 million English Wikipedia articles multiplied by number of paragraphs in each) and $k$ is usually small, such as 1-5. We used two distinct knowledge bases in our experiments: English Wikipedia and Wikidata. The English Wikipedia articles is parsed into a structured format to extract the content of each article. Articles are further divided into logical units, primarily paragraphs. Each paragraph is treated as an independent context unit for which questions are generated. A unique hash is computed for every paragraph which acts as the key to locate it in the original article. This ensures that when a question is retrieved, the associated original context can be found.

The Wikipedia content is primarily in Wikitext markup. It can also be rendered to HTML. But for the purpose of our system, we prepared plain text version of each passage so that it is comprehensible for an LLM. We also removed reference numbers that appear in usual plain text format. 

Let $\mathcal{D} = \{d_1, d_2, ..., d_M\}$ represent the knowledge base consisting of $M$ content units.

For each content unit $d_i \in \mathcal{D}$, a set of questions is generated by an LLM, denoted as $Q_i = \{q_{i1}, q_{i2}, ..., q_{in_i}\}$, where $n_i$ is the number of questions generated for $d_i$. The prompt used for the LLM is given in Appendix \ref{sec:llm-prompt-wikipedia-questions}. The input to the prompt not only contains the passage text, but contextual information like article title, section titles to resolve coreferences. The prompt also uses few shot prompting to get machine readable output. The concept of document expansion or enrichment by adding queries is common in information retrieval techniques\cite{nogueira-2019-documentexpansionqueryprediction}. Here we are not enhancing the document(passage), but uses the questions generated out of it.

Let $\mathcal{E}$ be the embedding function that maps a text input to a vector space, so $\mathcal{E}: \text{Text} \rightarrow \mathbb{R}^d$, where $d$ is the dimensionality of the embedding space. The embedding of a question $q_{ij}$ is denoted as $e_{ij} = \mathcal{E}(q_{ij})$, where $e_{ij} \in \mathbb{R}^d$. The set of all question embeddings is denoted as $\mathcal{V} = \{e_{11}, e_{12}, ..., e_{Mn_M}\}$.

We create an index $\mathcal{I}$ that maps the embeddings $e_{ij}$ to the corresponding content unit $d_i$ via the hash function $h$, such that $\mathcal{I}(e_{ij}) = h(d_i)$. The hashing function used is SHA-256 with 32 bytes length.

Let $q_u$ be the user's query. The embedding of the user's query is $e_u = \mathcal{E}(q_u)$.

We use cosine similarity $\text{sim}(e_u, e_{ij})$ to compare the user query embedding with each of the generated question embeddings. The best match is chosen using argmax function:

$$ j^* = \underset{j}{\text{argmax}} \ \text{sim}(e_u, e_{ij}) $$
where $j$ represents the set of all question embeddings in the question vector store.

The content corresponding to the most similar question embedding is retrieved using the hash function:
$$ d^* = \mathcal{I}(e_{j^*}) =  d_i \ \ \text{for some} \  e_{ij} $$

Let us illustrate this with the same example passage used in table \ref{similarity-score}. To make it more realistic, let use mimic users queries as incomplete sentences, often missing question words and occasional spelling mistakes. See table \ref{tab:example_retrieval}. As Generated questions are mapped to a passage that an answer it, the effective retrieval similarity for all user queries are same as the Similarity Score. 
Table \ref{tab:example_retrieval} illustrates the effectiveness of our approach in handling real world queries and accurately matching the passages.

\begin{table*}[h!]
\begin{tabularx}{\linewidth}{XXl}
        \toprule
        \textbf{User Query} & \textbf{Most Similar Generated Question} & \textbf{Similarity Score} \\
        \midrule
        "Obama's birthplace?" & "Where was Obama born?" & 0.91 \\
        \midrule
        "France nuclear energy percentage?" & "What percentage of France's electricity is nuclear?" & 0.92 \\
        \midrule
       "How many people died in chernobyl accident" & "How many people died in chernobyl disaster" & 0.97 \\
         \midrule
       "How many people died in chernobyl" & "How many people died in chernobyl disaster" & 0.97 \\
         \midrule
       "Deaths chernobyl accident" & "How many people died in chernobyl disaster" & 0.90 \\
        \midrule
       "Mayor of paris" & "Who is the current mayor of paris?" & 0.93 \\
          \midrule
       "longest river in Africa" & "Which is the longest river in Africa?" & 0.96\\
          \midrule
       "length of Nile" & "What is the total length of Nile river?" & 0.93
       
    \end{tabularx}
    \caption{Example of Question-to-Question Retrieval with Cosine Similarity Scores. Note the typos and omission of question words in user query to mimic real world search scenario}
    \label{tab:example_retrieval}
\end{table*}

The context granularity can be refined from paragraph to sentence level through Jaccard similarity matching between the query and individual sentences. When no strong sentence-level match is found, the system defaults to paragraph-level context.

\begin{figure*}
    \centering
    \includegraphics[width=1\linewidth]{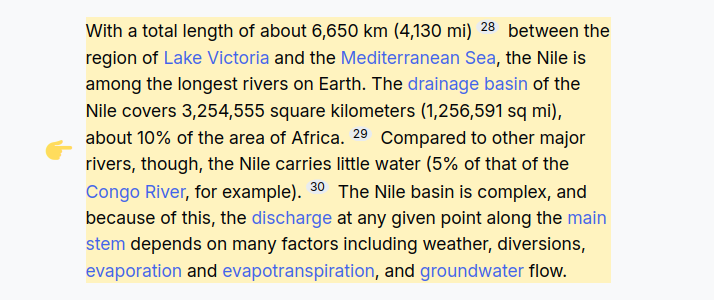}
    \caption{Screenshot from a prototype showing the answer presentation using our approach for the question "length of Nile". Upon entering the query, page navigates to Nile article, and scrolls to this paragraph and then highlights it.}
    \label{fig:enter-label}
\end{figure*}
\subsection{QA on Wikidata}

To incorporate Wikidata, we needed to extract factual information from the knowledge base in a format that is suitable for the question generation task.  We wrote a SPARQL query(See Listing \ref{wikidata-sparql}) that retrieves all subjects, predicates, and objects (triples) associated with each Wikidata entity (QID). The query is designed to be comprehensive in extracting fact triples in a human-readable textual format (e.g., “India: inception: 15 August 1947” or "India: Prime Minister: Narendra Modi (2014-current)").
The conversion of triples to text for querying on structured content was originally discussed in UniK-QA system\cite{oguz-etal-2022-unik}. A unique hash of QID and PID pair is calculated for linking the extracted content with generated questions. 

Let $Q$ be a Wikidata item (e.g., Q668 for India), and let $\mathcal{S}_Q$ be the set of statements associated with item $Q$. Each statement $s \in \mathcal{S}_Q$ can be represented as a tuple $s = (p, v, \mathcal{q})$, where:
\begin{itemize}
    \item $p$ is the property (PID).
    \item $v$ is the value.
    \item $\mathcal{q}$ is the set of qualifiers (optional).
\end{itemize}

Let $\mathcal{T}$ be the function that maps a statement $s$ to a textual representation $t$.
$$\mathcal{T}: s \rightarrow t$$
The function $\mathcal{T}$ involves:
\begin{itemize}
    \item Label retrieval for QIDs and PIDs.
    \item Formatting of dates and times into human-readable text.
    \item Concatenating subject, predicate, object and qualifiers.
     \item For simple cases,  $t$ will be in format of ``label(Q): label(p): value``.
     \item For statements with qualifiers, the textual representation will be: ``label(Q): label(p): value (label($q_1$): $v_1$, label($q_2$): $v_2$, ...)``, where $q_i$ is a qualifier and $v_i$ is the value of qualifier.
\end{itemize}
The set of all text triples generated from a given QID is denoted as $T_Q = \{t_1, t_2, ..., t_n \}$, where $n$ is the number of statements for item $Q$.

For each textual triple $t_i \in T_Q$, we generate a set of questions using an LLM, denoted by $Q_i = \{q_{i1}, q_{i2}, ..., q_{in_i} \}$. This process can be represented as:
$$Q_i = LLM(t_i)$$
Where $n_i$ is the number of questions generated for $t_i$. The prompt used for the LLM is given in Appendix \ref{sec:llm-prompt-wikipedia-questions}

The same embedding function $\mathcal{E}$ is used:
$$\mathcal{E}: \text{Text} \rightarrow \mathbb{R}^d$$
The embedding of a generated question $q_{ij}$ is denoted as $e_{ij} = \mathcal{E}(q_{ij})$. The set of all question embeddings is denoted as $\mathcal{V} = \{e_{11}, e_{12}, ..., e_{Mn_M}\}$, where $M$ is the total number of text triples across all items and $n_i$ is the number of questions for each triple.
 
We create an index $\mathcal{I}$ that maps the embeddings $e_{ij}$ to the corresponding source triple $t_i$ via the hash function $h$, such that $\mathcal{I}(e_{ij}) = h(t_i)$.

The rest of the steps are the same as the general approach:
\begin{itemize}
 \item Let $q_u$ be the user's query, the embedding is $e_u = \mathcal{E}(q_u)$.
    \item  Similarity is calculated as:
$$ j^* = \underset{j}{\text{argmax}} \ \text{sim}(e_u, e_{ij}) $$
 \item  The corresponding triple is retrieved:
$$ t^* = \mathcal{I}(e_{j^*}) =  t_i \ \ \text{for some} \  e_{ij} $$
\end{itemize}

Let's consider a few example Wikidata triplets related to "India" in the Subject: Predicate: Object format:
\begin{itemize}
     \item Subject: Q668 (India)
     \item Predicate: P571 (inception)
     \item Object: 1947-08-15T00:00:00Z
  \end{itemize}
The text representation for this is   ``India: Inception: 15 August 1947''. Similarly, we can have textual representations like:

\begin{itemize}
    \item India: Inception: 15 August 1947
    \item India: Prime Minister: Narendra Modi (2014-current)
    \item India: Life expectancy: 62 (1999)
    \item India: Flag: \url{https://commons.wikimedia.org/wiki/File:Flag_of_India.svg}
    \item India: Capital: New Delhi
\end{itemize}

Here are some example questions an LLM might generate for those Wikidata triplets:

\begin{itemize}
    \item When was India founded?
    \item When did India become independent?
    \item Who is the current prime minister of India?
    \item Who was the prime minister of India in 2020?
    \item What was the life expectancy in India in 1999?
    \item What is the average life span in India around the year 1999?
    \item Show me the flag of India.
    \item What does the flag of India look like?
    \item What is the capital of India?
    \item Where is the capital of India located?
\end{itemize}

See \ref{sec:question-matching-wikidata} for example comparison and similarity score.

Multimodal content (images, videos, 3D models) becomes searchable through question-based matching of their associated metadata. For instance, Wikidata triples like "Q243:P4896:[filename]" can be transformed into text ("Eiffel Tower: 3D Model: [filename]") and indexed with questions like "What is the 3d model of Eiffel Tower?"—enabling matches with user queries such as "Show me Eiffel tower 3d model." Similarly, Q140:P51:[audio file name] facilitates answers to queries like "How does the lion roar?"


\section{Discussion}
\subsection{Practical Considerations}

The increased vector store size—approximately tenfold due to question-based indexing rather than passage-based—remains manageable given modern vector databases' capability to efficiently search billions of records. While Wikipedia's frequent updates necessitate question regeneration, hash-based tracking enables selective re-indexing of modified passages only.
The experimental implementation indexed under 1,000 Wikipedia articles, utilizing llama-3.1-8b-instruct-awq for question generation and baai/bge-small-en-v1.5 (384-dimensional vectors) for embeddings, with processing conducted as a background task.

\subsection{Advantages}
Our approach offers several distinct advantages over traditional RAG systems:
\begin{itemize}
    \item Hallucination-Free Responses: The most important advantage is that the response is hallucination-free, since we directly retrieve the original content instead of relying on generated answers.
    \item High Precision Retrieval: Comparing questions with questions leads to highly accurate results with very high cosine similarity scores.
    \item Fast Retrieval Time: Due to the absence of LLM calls during inference time, query time is significantly lower than that of RAG based systems. This also implies significant cost savings. 
    \item Multimodality: Retrieval of images, audio and other forms of data are handled seamlessly because we retrieve facts as structured text and treat them uniformly with other textual context.
\end{itemize}

\subsection{Limitations}
The system focuses on direct fact retrieval (extractive). Answering more complex questions that require multi-hop reasoning, aggregation, or synthesis might require further advancements. For questions that need more than just one paragraph or fact, the system might fall short. While the diversity of generated questions is high, it might be possible to improve question generation coverage to improve system performance for complex questions. LLMs are efficient only in a small set of languages compared to 300+ languages where Wikipedia exist. We have not evaluated the effectiveness of this approach in languages other than English.

\section{Conclusion}

In this paper, we introduced a novel question-to-question retrieval approach for open-domain question answering over Wikipedia and Wikidata, which achieves high precision and eliminates the risk of hallucination by avoiding the traditional generation step. Our approach leverages an instruction-tuned LLM to generate comprehensive sets of questions for each content unit in the knowledge base, which are then embedded and indexed for efficient retrieval. This method leads to cosine similarity scores consistently above 0.9 for relevant question pairs, demonstrating a high degree of alignment between user queries and the indexed content. By directly retrieving the original text and structured data based on the best matching generated questions, our system offers a fast, scalable, and reliable alternative to conventional RAG pipelines, with the added benefit of supporting pseudo-multimodal question answering.

\section*{Acknowledgments}

This paper benefited from the valuable contributions of Isaac Johnson and Xiao Xiao of the Wikimedia Foundation.

\bibliography{references}
 
\clearpage
\appendix
\section{LLM Prompts}

\subsection{LLM prompt to generate all possible questions for a given passage}

You are an expert question generator tasked with creating simple, natural-language questions
from a given Wikipedia article that people might typically search on the internet
based on a provided text passage.

The input will include:
\begin{itemize}
\item Article Title
\item Section Title
\item Paragraph Text
\end{itemize}

\subsubsection*{Guidelines for Question Generation}

    Use the article and section titles to resolve any ambiguous references in the text
    Create questions that can be directly answered by the text
    Prioritize who, what, where, when, and how questions
    Ensure questions are simple, clear and concise mimicking common search engine query patterns
    Avoid yes/no questions unless the answer is explicitly stated in the text
    Avoid generating questions that require external knowledge not present in the text
    Avoid generating speculative or opinion-based questions
    Avoid very long questions that may be difficult to understand
    Include questions about: \begin{itemize} \item Key concepts \item Specific details \item Important processes \item Significant events or characteristics \item Dates, places, and people \item Questions that can be answered by current subject \end{itemize}

\section*{Output Format}
\begin{itemize}
\item Provide a bullet list of questions
\item Each question should be a single, complete interrogative sentence
\end{itemize}

\section*{Example Processing}
\textbf{Input:}

Article Title: Barack Obama
Section Title: Early Life and Education

Obama was born in Honolulu, Hawaii.
He graduated from Columbia University in 1983 with a Bachelor of Arts degree in political science and later worked as a community organizer in Chicago.
In 1988, Obama enrolled in Harvard Law School, where he was the first black president of the Harvard Law Review.
He became a civil rights attorney and an academic, teaching constitutional law at the University of Chicago Law School from 1992 to 2004.
In 1996, Obama was elected to represent the 13th district in the Illinois Senate, a position he held until 2004, when he successfully ran for the U.S. Senate.
In the 2008 presidential election, after a close primary campaign against Hillary Clinton, he was nominated by the Democratic Party for president.
Obama selected Joe Biden as his running mate and defeated Republican nominee John McCain.

\textbf{Expected Output:}

\begin{itemize}
\item Where was Barack Obama born?
\item What state was Obama born in?
\item Which university did Obama graduate from?
\item What year did Obama graduate?
\item What was Obama's major in college?
\item What did Obama do after graduating from Columbia?
\item Where did Obama work as a community organizer?
\item When did Obama enroll in Harvard Law School?
\item Who is the first black president of Harvard Law Review?
\item From what years did Obama teach at University of Chicago Law School?
\item When was Obama first elected to the Illinois Senate?
\item When did Obama run for U.S. Senate?
\item Who did Obama compete against in the Democratic primary?
\item Who was Obama's running mate in the 2008 presidential election?
\item Who did Obama defeat in the 2008 presidential election?
\item Who defeated John McCain in the 2008 presidential election?
\item What political party nominated Obama for president?
\end{itemize}

\subsection{LLM prompt to generate all possible questions for a Wikidata statement}

You are a specialized question generation system. Your task is to convert knowledge triplets into natural questions. Each triplet follows the format:

[Subject] : [Predicate] : [Object]

\section*{Guidelines for question generation:}

    Transform the predicate into an appropriate question word:
    \begin{itemize}
    \item "founded by" -> "Who founded"
    \item "located in" -> "Where is"
    \item "born in" -> "When was"
    \item "invented" -> "What did"
    \item "composed" -> "Who composed"
    \end{itemize}

    Question formation rules:
    \begin{itemize}
    \item Start with the appropriate question word (Who, What, Where, When, How)
    \item Place the subject appropriately in the question
    \item End all questions with a question mark
    \item Maintain proper grammatical structure
    \item Preserve proper nouns and capitalization
    \item Remove the predicate's passive voice if present ("founded by" → "Who founded")
    \end{itemize}

    Handle special cases:
    \begin{itemize}
    \item Multiple objects: Generate separate questions for each object
    \item Complex predicates: Break down into simpler components
    \item Dates: Use "When" for temporal relations
    \item Locations: Use "Where" for spatial relations
    \end{itemize}

\section*{Examples:}

\textbf{Input:} "San Francisco : founded by : José Joaquín Moraga, Francisco Palóu"
\\
\textbf{Output:}
\begin{itemize}
\item Who founded San Francisco?
\end{itemize}

\textbf{Input:} "Eiffel Tower : located in : Paris"\\
\textbf{Output:}
\begin{itemize}
\item Where is the Eiffel Tower?
\end{itemize}

\textbf{Input:} "JavaScript : created by : Brendan Eich"\\
\textbf{Output:}
\begin{itemize}
\item Who created JavaScript?
\end{itemize}

\textbf{Input:} "Theory of Relativity : developed in : 1905"\\
\textbf{Output:}
\begin{itemize}
\item When was the Theory of Relativity developed?
\end{itemize}

Always ensure questions are:
\begin{itemize}
\item Grammatically correct
\item Natural sounding
\item Unambiguous
\item Focused on a single piece of information
\item Answerable using the information in the original triplet
\end{itemize}

\section{SPARQL Query for getting all statements for a given Qitem}

\lstinputlisting[label={wikidata-sparql}, float=*t, caption={SPARQL Query for getting all statements for a given Qitem},language=SPARQL, showspaces=false,basicstyle=\ttfamily\small]{qitem.sparql}

\section{Example question-question matching for Wikidata}
\label{sec:question-matching-wikidata}
\begin{table*}
       \begin{tabularx}{\linewidth}{XXc}
        \toprule
        \textbf{User Query} & \textbf{Most Similar Generated Question} & \textbf{Cosine Similarity Score} \\
        \midrule
        "When was India established?" & "When was India founded?" & 0.97 \\
         \midrule
        "Who leads India now?" & "Who is the current prime minister of India?" & 0.94 \\
        \midrule
       "What was life expectancy in India back then?" & "What was the life expectancy in India in 1999?" & 0.92 \\
        \midrule
        "Show Indian flag" & "Show me the flag of India." & 0.95 \\
        \midrule
        "India's capital city" & "What is the capital of India?" & 0.96 \\
        \midrule
         "Tell me the time of Indian independence" & "When did India become independent?" & 0.95 \\
         \midrule
           "Who is the PM of India now" & "Who is the current prime minister of India?" & 0.96 \\
           \midrule
        "What is the flag of India like?" & "What does the flag of India look like?" & 0.92 \\
         \midrule
         "Where is the capital of India located" & "Where is the capital of India located?" & 0.97 \\
        \midrule
           "Average life span of India?" & "What is the average life span in India around the year 1999?" & 0.93 \\
        \bottomrule
    \end{tabularx}
    \caption{Example of Wikidata Question-to-Question Retrieval with Cosine Similarity Scores}
    \label{tab:wikidata_retrieval}
\end{table*}

\end{document}